\documentclass[lettersize,journal]{IEEEtran}
\usepackage{amsmath,amsfonts}
\usepackage{algorithmic}
\usepackage{algorithm}
\usepackage{array}
\usepackage[caption=false,font=normalsize,labelfont=sf,textfont=sf]{subfig}
\usepackage{textcomp}
\usepackage{stfloats}
\usepackage{url}
\usepackage{verbatim}
\usepackage{graphicx}
\usepackage{cite}
\usepackage{color}
\begin{document}

\title{A Framework for Predicting the Impact of Game Balance Changes through Meta Discovery}

\author {
    Akash Saravanan, Matthew Guzdial \\
    University of Alberta\\
    asaravan@ualberta.ca, guzdial@ualberta.ca
}

\markboth{IEEE Transactions on Games, Special Issue on Human-Centred AI in Game Evaluation}%
{Special Issue on Human-Centred AI in Game Evaluation}


\maketitle

\begin{abstract}
A metagame is a collection of knowledge that goes beyond the rules of a game. In competitive, team-based games like Pok\'emon or League of Legends, it refers to the set of current dominant characters and/or strategies within the player base. Developer changes to the balance of the game can have drastic and unforeseen consequences on these sets of meta characters. A framework for predicting the impact of balance changes could aid developers in making more informed balance decisions. In this paper we present such a Meta Discovery framework, leveraging Reinforcement Learning for automated testing of balance changes. Our results demonstrate the ability to predict the outcome of balance changes in Pok\'emon Showdown, a collection of competitive Pok\'emon tiers, with high accuracy.
\end{abstract}


\section{Introduction} \label{sec: balancing_game_characters}

Game balance in competitive multiplayer games is a constant process. 
In character-based games like Pokémon or League of Legends there are regular patches adding new characters or adjusting existing ones. 
This impacts the metagame (or meta), which is the set of current dominant characters or strategies within the player base. 
The metagame can be constantly changing, with players continually inventing new dominant strategies to counter the current metagame. In time, these new strategies could become the prevalent meta. This process then repeats, allowing for the metagame to achieve a level of self-balancing. 
However, in other scenarios, there could be one clear dominant strategy that is superior to all others. In addition, the metagame could stagnate or reach a state where the dominant strategies are not fun to play against.
This negatively impacts player experience and can lead to players abandoning a game.
Developers need to be careful when they attempt to make changes to alter the metagame. There have been instances of developer changes causing major, unintended shifts in the metagame such as the infamous Juggernaut patch of League of Legends \cite{metagame-shifts} and the creation of the Anything Goes tier in competitive Pok\'emon \cite{anything-goes-creation}. Thus the meta that arises from a balance change may not necessarily be the meta that the developers envisioned.

When balancing games, many developers consider the combination of a character's winrate and their pickrate \cite{rito-balance-framework}. The pickrate is the percentage of matches a character is picked in while the winrate is the percentage of matches that the character actually wins. We note that while this information offers value in identifying areas that may need changes, it does not specify what changes to make. In practice, developers employ their expertise, domain knowledge, and regular testing to make balance changes. We note that we do not attempt to automate this process, merely give additional insight into the likely impact of balance changes.
Outside of developers, there exist automated balancing approaches \cite{zohaib2018dynamic}. 
However, these tend to focus on balancing single-player experiences, not multiplayer games. 
In addition, they make changes directly rather than predicting the impact of potential changes. 

\begin{figure*}[t]
    \centering
    \includegraphics[width=6.3in]{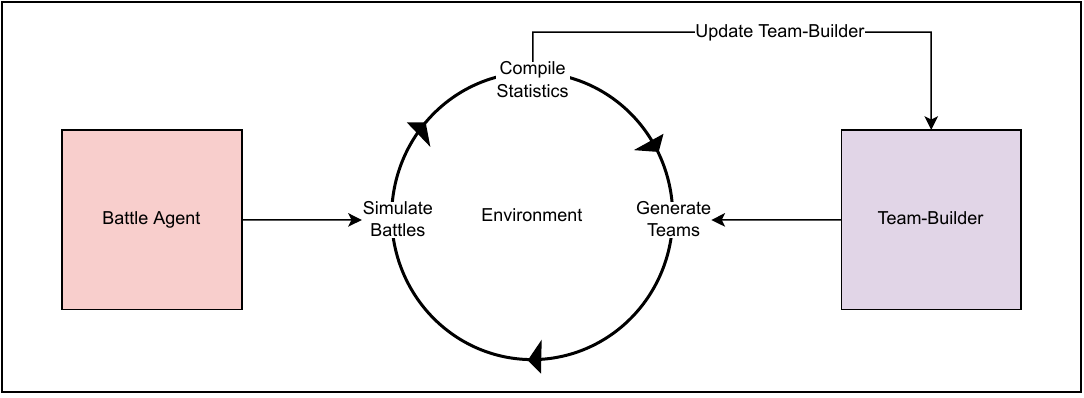}
    \caption{The three main components of our meta discovery framework. The Battle Agent plays matches while the Team Builder generates teams (using a mix of metagame statistics and domain knowledge) for the Battle Agent to use in matches. The simulator Environment is responsible for simulating battles using the Battle Agent with the generated teams from the Team Builder and tracking the overall statistics of the system.}
    \label{fig:system_overview}
\end{figure*}

In this paper, we propose a Meta Discovery Framework for team-based, character-dependent competitive games, as illustrated in Figure \ref{fig:system_overview}. 
By team-based we indicate that matches take place between two teams.
By character-dependent we indicate that the games rely on in-game characters, who make up the aforementioned teams. 
This framework consists of three components - a battle agent to play matches, a team-builder for approximating new meta teams, and an environment to simulate matches after a balance change. 
Using this framework, we attempt to discover the impact of changes to the metagame such as the addition, removal, or adjustment of characters. 
We name this task \textbf{ABC-Meta}, or Analyzing Balance Changes on the Metagame.
In essence, we evaluate game balance by analyzing the results of changes made to it.
We analyze the ABC-Meta task on Pok\'emon Showdown, a competitive Pok\'emon environment and draw on historic balance changes to test our approach. 
Our results demonstrate that this approach can predict the impact of balance changes with high accuracy. 
Our hope is that this framework can help developers to better understand the potential impact of changes made to characters before deploying them into the game's live environment. 

Our contributions in this paper are as follows: 
\begin{enumerate}
    \item The development of a Meta Discovery Framework and instantiations of it with Reinforcement Learning and heuristic agents.
    \item The proposal of the ABC-Meta task as a new games research challenge.
    \item The proposal of a related Blank Slate Discovery of Metagames or \textbf{BSD-Meta} task.
    \item Results in Pok\'emon Showdown on historic data, supporting the utility of the framework and the feasibility of our identified tasks.
\end{enumerate}

\section{Background}

\subsection{Pok\'emon}\label{related_bgpixel_pokemon}
The Pok\'emon video game series are Role-Playing Games (RPGs) that involve players capturing the titular Pok\'emon creatures and fighting battles with them. With every new release, the developers introduce a significant number of new Pok\'emon, new abilities, items, mechanics and more. At the time of this writing, there are over 1000 Pok\'emon creatures. Thus, it is very difficult to predict how a new Pok\'emon will interact with the existing ones, making it valuable to be able to predict the impact of any changes to the metagame.

\subsection{Pok\'emon Showdown}\label{meta_experimental}
To evaluate our system, we implement it in the Pok\'emon video game series, specifically on Pok\'emon Showdown. 
Pok\'emon Showdown is an online Pok\'emon battle simulator with millions of monthly matches \cite{smogon-stats}.
It has an extensive, complex, and publicly documented metagame. 
We use Pok\'emon over other games for two reasons - the battles are generally much faster and it has a simple, easy to access API. 
Battle speed is important as it allows us to simulate a large number of matches in a relatively short amount of time, thus allowing us to experiment with different techniques for meta discovery. 
Other games such as League of Legends and Dota 2 take much longer for matches (around 30 minutes on average) while Pok\'emon battles usually take less than 5 minutes. 
We elect to use Pok\'emon Showdown in particular over something like the VGC format \cite{vgc-ai} due to its popularity and easily available metagame data. 

\subsection{Attributes of Competitive Pok\'emon Battles}\label{related_bgpixel_pokemon_battleattrs} 
In terms of competitive battling, each Pok\'emon has several stats and attributes to take into account. While a complete understanding is not necessary to understand this work, they are important aspects of competitive Pok\'emon battles and serve to illustrate the complexity and interlinked nature of these battles. Every Pok\'emon has a level (1-100), 6 base stats (Hit Points [HP], attack, defense, special attack, special defense, and speed) each in a range of 1-255, 1 to 4 moves (from over 800 options across 3 classes), one or two types (from 18 choices), Effort Values (EVs) ranging from 0 to 255 and Individual Values (IVs) ranging from 0-31 for each of the 6 base stats, a nature, 2 to 3 passive abilities (from over 250 options) and up to 1 item (from over 500 options). Every move also has its own statistics such as Accuracy and Base Power, as well as one of the 18 types. There is also a 1 in 24 base chance of a move being a critical hit, which deals increased damage. Each Pok\'emon type has a set of other types that it is strong against, weak against, resistant to, and immune to. For example, fire-type is weak to water-type and strong against grass-type. Further, Pok\'emon have a Same Type Attack Bonus (STAB) which causes them to deal more damage with moves that match their type. A Pok\'emon's overall stats are calculated using a combination of its base stats, effort values, individual values, level and nature. Some moves, abilities and items may also inflict one of 6 permanent status effects or over 150 temporary status effects each of which influences a Pok\'emon's attributes. 
The Pok\'emon's overall stats, in addition with its type, a move's base power, STAB, ability, item, stat changes and status effects determine the amount of damage a Pok\'emon can do with a single move. 

\subsection{Pok\'emon Battles}\label{related_bgpixel_pokemon_battle} 
In a standard battle each player has a team of up to 6 Pok\'emon. 
In competitive Pok\'emon matches, all Pok\'emon are usually level 100, utilize the maximum amount of EVs and IVs, have 4 moves, and equip an item. At the start of the battle, each player chooses one of their 6 Pok\'emon to send out. In some battle formats this choice will always be the first Pok\'emon in the roster and in others it can be random. Battles proceed in a turn-based fashion where both players perform an action simultaneously, the results are calculated and the players once again take an action. On a player's turn they may either choose one of the 4 moves that a Pok\'emon knows or they may switch to one of the available Pok\'emon on their team. If a player's Pok\'emon has fainted (its HP has been reduced to 0), the Pok\'emon will no longer be available for the battle. In this case, the player must switch to another Pok\'emon on their team. This is a free action, that is, the opponent does not take an action until a new Pok\'emon has been sent in (the player's next turn). This continues until a player forfeits or all 6 of a player's Pok\'emon have fainted (HP reduced to 0). 
There are also alternate formats such as the Doubles format, where each player fields two Pok\'emon at a time. However this format is outside the scope of this paper and so we do not discuss it in further detail.

The team-building aspect of Pok\'emon adds in another layer of complexity. As seen in the previous section, just considering all the attributes of a single Pok\'emon is a significant challenge. When teams come into play, this problem repeats six times. In addition, other factors need to be taken into account. These include how well each Pok\'emon synergizes with the team, major or minor weaknesses in the team against certain Pok\'emon, and the likelihood of such threats appearing in a battle, and others. Additionally, there are several distinct Pok\'emon archetypes and overall play styles \cite{showdown-competition}. Thus this team-building aspect of Pok\'emon is an entire research topic of its own \cite{technical-machine,fsai}.

\subsection{The Competitive Metagame of Pok\'emon}\label{related_bgpixel_pokemon_comp}

Pok\'emon Showdown is a popular online simulator for Pok\'emon battles, which is sponsored by Smogon \cite{smogon}, a fan-run competitive community. There are several different battle formats or ``tiers" used in Pok\'emon Showdown. Each of these tiers have their own metagame with varying strategies and characters. The most popular format according to publicly available statistics is the random battle format where players do not build their own team \cite{smogon-stats}. Instead, both players use a pseudo-randomly generated team. However, this format does not have a true metagame due to this pseudo-random generation. In fact, the generation itself depends on the existing metagame. Most other tiers also have several clauses put into place for both game balance and fun. These include a ``species clause" which prevents multiple of the same Pok\'emon species from being used along with several others that ban particular moves or abilities. In addition, each tier also has its own particular banlist of Pok\'emon, moves, abilities or items. The tiering system, bans and clauses are all created and maintained by Smogon.

The most popular competitive tier is the OverUsed (OU) tier. Below OU is the UnderUsed (UU) tier. This continues on with the NeverUsed (NU), RarelyUsed (RU), and PU (no full form) tiers. In each case, all Pok\'emon with a certain weighted usage rate are classified as being a Pok\'emon of that tier. A Pok\'emon from a higher tier is banned from usage in lower tiers, although lower tier Pok\'emon can be used in any higher tier. 
Each tier has an associated BanList (BL) which consists of Pok\'emon that are banned from that particular tier, but do not have enough usage to be classified into a higher tier. For instance, the UnderUsed BanList (UUBL) consists of Pok\'emon banned from UU, but under the required usage rate to be classified as OU. The one exception to this is the OU tier itself. All Pok\'emon banned from OU are placed in an entirely different tier - the Ubers tier. There are several other special tiers and cases, but these are the most popular ones. For simplicity, in this research we only focus on OU, UU, NU, and PU. 

\section{Related Work}\label{related_balance}

We now discuss prior research related to our work.  Specifically, in subsection \ref{related_balance_teams}, we examine systems to generate teams. In subsection \ref{related_balance_discovery} we cover work on the impact of balance changes in games as well as approaches surrounding the idea of meta discovery.

\subsection{Team-Building}\label{related_balance_teams}
Team-building is essential in team-based competitive games. When attempting to simulate the metagame, the teams that are generated directly form the meta. Thus team-building is an essential component of meta discovery. Several approaches have analyzed this aspect of competitive games. 
Summerville, Cook and Steenhuisen \cite{Summerville_Cook_Steenhuisen_2021} and Hong, Lee and Yang \cite{league-champ-reco} developed systems to predict the heroes picked during the drafting phase in Dota 2 and League of Legends, respectively. Both cases can be thought of as a team-building process using the current metagame alongside historical data.
Both Crane et. al \cite{pogo-counter} and Oliveira et. al \cite{pogo-optimization} studied team-building in Pok\'emon GO. The former tested several team generation systems to generate teams capable of regularly winning in an analyzed metagame while the latter tested several optimization algorithms to counter a given team. However, due to Pok\'emon GO being distinct from, and far simpler than Pok\'emon Showdown their system is not applicable to our work. 
Focusing on Pok\'emon Showdown in particular, Rejim \cite{rejim-time-efficient} developed a collaborative system to recommend additions to a team based on user specified inputs and usage statistics. Future Sight AI \cite{fsai} also generates teams based on usage statistics, but has no publicly available algorithm or implementation. Technical Machine \cite{technical-machine} either steals teams from opponents it loses to or uses a weighted random selection based on usage statistics. Our team-building framework uses techniques from Rejim and Technical Machine in order to dynamically generate teams based on the metagame state.

\subsection{Metagame Discovery \& Game Balance}\label{related_balance_discovery}

Kokkinakis et. al \cite{kokkinakis2021} studies the evolution of the concept of a metagame and how it affects game balance. On the application side, several instances of prior work have studied the modification of game rules and their resulting impact. These can be treated as changes to gameplay mechanics instead of changes to characters. 
One such work \cite{tomavsev2020assessing} focused on the game of chess, defining several alternate rulesets to the game and analyzing the resultant game after achieving near-optimal performance using AlphaZero. Though effective for chess, competitive Pok\'emon battling has no game playing agent that approaches this level of expertise. 
Jaffe et al. \cite{jaffe-restricted-play} restricted particular actions in a simple competitive game to evaluate the effects on game balance. However, their tool was built for perfect-information games, while Pok\'emon is an imperfect-information game. 
Zook, Fruchter and Riedl \cite{zook-automated-playtesting} balanced parameters of a game by combining an automated playtester with an active learning framework. However, they use a flat set of parameters and acknowledge the need for alternative techniques for more complex scenarios. 
Silva et al. \cite{hearthstone-meta} used an evolutionary algorithm to find a set of balance changes to Hearthstone cards such that decks approach a 50\% winrate. However, they use a small set of predefined decks which do not adapt to balance changes, and thus, do not use a team-building component.
Finally, Hernandez et. al \cite{hernandez-autobalancing} developed a system to balance competitive games by finding a parameterization of the optimal metagame, but required game balance to be represented as a parameterizable graph.

\section{Battle Agents}\label{meta_battle}

In this work, we draw on two different battle agents for comparison purposes, a heuristic agent and a Reinforcement Learning (RL) agent. 
Both are used in the framework depicted in Figure \ref{fig:system_overview} to test the impact of balance changes. 

We preface this section with the statement that our objective is not to create a state-of-the-art agent. Although this is an interesting research problem on its own, our goal is to only have an agent capable of approximating an average human at the game. This is because they have the most influence over the metagame, given that the meta is a measure of popularity and the majority of players in competitive online games are of average skill. However, an argument can be made that the metagame may vary based on a player's skill level (and thus matchmaking rating) \cite{rito-balance-framework}. For instance, if we have a state-of-the-art agent, the meta might reflect more high-level play which may not necessarily be true at lower skill levels. Thus, in an ideal scenario, having a swarm of agents of varying skill level would result in a truer meta. This is out of scope for this paper and we leave an exploration of this for future work.

We note that while the rest of this section details the specifics of setting up agents for Pok\'emon Showdown, our framework itself only requires an agent that can play like an average human. Thus future work does not need to specifically follow this approach.

We use the Random Battle format, specifically ``gen8randombattle", for training agents. This format involves two randomly generated teams of six Pok\'emon in a singles battle format. That is, each player sends out one Pok\'emon at a time and they battle until a player either forfeits or has no remaining Pok\'emon. We describe this system in more detail in Section \ref{related_bgpixel_pokemon_battle}. We note that the random battle format is different from the standard Smogon tiers that we analyze for meta discovery. We train our agents on this format for two reasons. First, this format does not require us to provide teams, instead pseudo-randomly generating them based on hand-crafted rules. In contrast, if we used the standard Smogon tiers, we would have to compile teams ourselves or create our own team-building algorithm. Second, the pseudo-random nature of the teams allows for more generalization in the reinforcement learning model since it offers a wider variety of teams to the model. We do note that some attributes of the generated team are dependent on the current meta. Thus we cannot use the team generation system used by gen8randombattle or other meta-dependent approaches as it would bias the results.

In order to train these agents, we make use of the poke-env API by Haris Sahovic \cite{poke-env}, which is a Python interface to create bots for Pok\'emon Showdown. All code used in this paper is publicly available in the linked Github repository\footnote{\url{https://github.com/akashsara/meta-discovery}}.

\subsection{Action Space}\label{meta_battle_action}
Our reinforcement learning agent has a discrete action space of size 22. The first four actions correspond to each of the four moves that a Pok\'emon can have. The next 12 actions correspond to Z-Move variations, Mega Evolution variations, and Dynamax variations of the same 4 moves. Each of these are once-per battle gameplay mechanics and are associated with the first four actions. The final six actions correspond to the ability to swap out the current Pok\'emon for the Pok\'emon in the corresponding position within the team. This does mean that at all times, at least one of these 6 actions will be illegal as one Pok\'emon is always on the field. Similarly, it is not possible to switch to a fainted Pok\'emon nor is it possible to use a move if the current Pok\'emon has fainted. We perform action masking, thus preventing such illegal actions from being taken. At every step, actions are passed to the environment as an integer corresponding to one of the 22 possible actions.

\subsection{State Representation}\label{meta_battle_state}

Our state representation is based on the default state representation available in the poke-env API. It is of size 10, with the first 4 elements corresponding to the base power of the Pok\'emon's 4 moves. We normalize this base power by dividing it by 100 to facilitate learning. The next 4 elements correspond to the type effectiveness of each move against the opponent's currently active Pok\'emon. The final two elements offer an indication of how many Pok\'emon are still alive on each team. We normalize this value by dividing it by 6 as a team can have a maximum of 6 Pok\'emon. There may be other aspects of a battle important for the state representation, but we use the default for simplicity and given that our goal is not to create a state-of-the-art agent.

\subsection{Reward Function}\label{meta_battle_reward}
Our reward function is a default provided by the poke-env API. It is, in essence, a weighted sum of the condition of the player's Pok\'emon (number of Pok\'emon and their current health), the opponent's Pok\'emon, and the state of the battle (win/loss). We use the weights recommended by the author of poke-env. Specifically, we weigh fainted Pok\'emon and overall HP features at 0.0125 and 0.1 respectively and give out a reward of 1 and -1 for victories and defeats respectively. The specific implementation of this function can be found in the poke-env library and we omit it from this work due to its complexity.

\subsection{Agents}\label{meta_battle_baseline}

As discussed above, we try two battle agents to evaluate our framework: a heuristic agent and a reinforcement learning agent. The heuristic agent uses hand-crafted rules and domain knowledge. This agent was developed by the authors of poke-env \cite{poke-env}. It is capable of estimating stats of opponents, determining the use of specific types of moves and the use of one-time battle mechanics, analyzing type effectiveness and STAB, and determining when to switch out to a different Pok\'emon. 

We additionally train a reinforcement learning agent using the PPO algorithm with Generalized Advantage Estimation due to its success in prior work \cite{huang-selfplay}.
This agent consists of two feedforward layers with 128, and 64 nodes respectively, with a standard ELU \cite{elu} activation function following each layer. We settled on this architecture through our initial experiments. In order to output both a policy and a value as required by the PPO algorithm, we utilize two output layers corresponding to the actor-head and the critic-head. The former is mapped to the 22 actions in the action space while the latter outputs a real numbered value for the given state.
We use a fixed random seed of $42$. The agent uses a discount factor of $0.95$, a Generalized Advantage Estimation Lambda value of $0.9$, a surrogate clipping parameter of $0.1$, a value function clipping parameter of $0.1$, and loss constants of $0.5$ and $0.002$. We use a batch size of $128$ and an Adam optimizer with a learning rate of $2e-4$. All battles occur via self-play, that is, the model faces itself. We perform $10,000$ steps per rollout, train for $10$ epochs after each rollout (for a total of $1,024,000$ steps). One step is a single turn in a battle. All hyperparameters and the training process are adapted from prior work \cite{huang-selfplay} with some empirical modifications.

\subsection{Agent Evaluation}\label{meta_agent_eval}
We perform two different evaluations of our model. The first is a qualifier used during our preliminary experiments - the agent's winrate against a completely random agent. The heuristic agent won $993$ of $1000$ games while the RL agent won $978$. Our second evaluation was to test the agents on the Pok\'emon Showdown ladder. This lets us test the agents against human players and use Pok\'emon Showdown's own metrics to determine if the agents meet our criteria of approximating an average human. One way to interpret this criteria would be to say an agent wins $50\%$ of the battles it plays. To this end, we look at each agent's GXE (Glicko X-Act Estimate) value \cite{smogon-ratings} after $100$ battles on the ladder. GXE measures the odds of a player winning a battle against a randomly selected opponent from the ladder. The heuristic agent achieved a GXE of $49.4\%$ while the RL agent reached a GXE of $42.4\%$. Thus the heuristic agent is closer to an average player while the RL agent is somewhat worse. However, we still use both agents for our future analyses to improve the breadth of our results.

\section{Team-Building}\label{meta_teambuilding}

The remaining part of our Meta Discovery Framework is the team-builder, which attempts to approximate meta teams given the simulated battles of the Battle Agent.
Since the overall meta is dependent on the most popular characters, the method through which teams are generated will influence the meta discovered through the framework. Thus we need to generate teams in a manner similar to that used by human players. This is true in the case of most games, not just Pok\'emon.
Our solution for this problem is inspired by prior work \cite{rejim-time-efficient,technical-machine,fsai} and uses a combination of domain knowledge (type synergies, base stats) and metagame knowledge (pickrates, winrates). Domain knowledge such as a Pok\'emon's type or stats are usually fixed and do not change with the meta. Metagame knowledge is constantly evolving as the metagame shifts. 

We analyze the team-building problem from the lens of both the Analyzing Balance Changes in a Metagame (ABC-Meta) task and the Blank Slate Discovery of the metagame (BSD-Meta) task. 
In the case of the ABC-Meta task, we have access to the true metagame before a balance change. In the case of the BSD-Meta task, we do not have access to the true metagame and thus have to start from scratch. 
In both cases, the domain knowledge used remains fixed while the metagame knowledge is continually updated throughout the simulation.

Similar to prior work \cite{rejim-time-efficient,technical-machine,fsai} we represent the team-building process sequentially, automatically picking one member of the team at each step. We call the team being built the \textbf{Current Team} and the set of available characters as the \textbf{Pool}. Initially the Current Team will be empty and at each step of the team-building process, one characters is added to it. 
In order to select a character to add to the team, we include several components using both domain and metagame knowledge. In the next section, we define these components.

Before building a team however, we first need to identify each individual Pok\'emon's moveset (all the aspects discussed in Section \ref{related_bgpixel_pokemon_battleattrs}). To obtain this, we scrape publicly available Smogon data that compiles such movesets based on usage statistics. We acknowledge that this information would not be available for the BSD-Meta task and would instead require an alternative method. We leave this for future work.

\subsection{Components}\label{meta_teambuilding_components}
We first consider three statistics that use current metagame knowledge. 
All the statistics we consider have a value ranging from 0.0 to 1.0 and are computed on a per-Pok\'emon basis in a vectorized manner.
The first statistic is the pickrate which refers to the fraction of teams in which a Pok\'emon has been picked. Since both teams in a battle can have the same Pok\'emon, the pickrate for a Pok\'emon is defined as the total number of picks divided by twice the number of battles.

\begin{equation}
    \text{Pickrate(X)} = \frac{\text{Num. Picks(X)}}{2 \times \text{Num. Battles}}
\end{equation}

Next, the winrate refers to the fraction of battles that a Pok\'emon wins. If a Pok\'emon has never been picked, the winrate is automatically set to 0 to avoid divide-by-zero errors.

\begin{equation}
    \text{Winrate(X)} = \frac{\text{Num. Wins(X)}}{{\text{Num. Picks(X)}}}
\end{equation}

Finally, the popularity as defined by Rejim \cite{rejim-time-efficient} is based on how frequently Pok\'emon are used together. We alter this to be a measure of how frequently a Pok\'emon wins when used alongside Pok\'emon already in the Current Team. This is used only when the Current Team has at least one Pok\'emon. The number of wins every Pok\'emon has when used with every other Pok\'emon is tracked in a popularity matrix. When team-building, this matrix is normalized.
For each Pok\'emon, we obtain the normalized value associated with every Pok\'emon in the Current Team. These values are then averaged to arrive at the overall popularity for that Pok\'emon. 

\begin{equation}
    \text{Popularity(X)} = \frac{\sum_{i \in \text{current team}}\text{PopMatrix[X][i]}}{|\text{current team}|}
\end{equation}

We now consider three statistics that utilize domain knowledge or a combination of domain knowledge with metagame knowledge. First, we use the Base Stat Total or BST, which is the normalized sum of a Pok\'emon's base stats. This value is effectively a constant as the BST cannot be changed by players. 

Next, we use two values based on type effectiveness in Pok\'emon, the Meta Type Value and the Type Value. We introduce the former while the latter is adapted from Rejim's work \cite{rejim-time-efficient}.
The former prioritizes Pok\'emon with types that are strong against the current meta while the latter prioritizes Pok\'emon with types that are strong against the types that can beat the Current Team. That is, the Type Value is higher for Pok\'emon that counter the counters to the Current Team. To obtain these values, we first pre-compute a matrix detailing the effectiveness of every type against every type. This type chart is assigned values of -2, -1, 0, and 1 for type immunities, resistances, neutralities and weaknesses respectively. Thus, given a type we can extract a type vector that details the effectiveness of all types against it. We extend this to a group of types by summing up the type vectors for the entire group. We then normalize this vector to obtain a value for each type and assign these values to Pok\'emon based on their type. In cases where Pok\'emon have two types, we take the average of these values.

To calculate the Meta Type Value, we compile the types of all the Pok\'emon in the current meta and calculate the value using the process described above. To calculate the Type Value, we first compile the types of all Pok\'emon in the Current Team. We then compile the group of types that are strong against the types in the current team and calculate the group of types that can beat that group. We pass this final group into the process described above. 

\subsection{Algorithms}\label{meta_teambuilding_algos}
To build a team, we compute a score for every Pok\'emon in the Pool. We then use this value to compute an approximation of the probability of a player picking each Pok\'emon. We then sample according to these probabilities to pick a Pok\'emon to add to the Current Team. 
We note that Pok\'emon that are banned are included in the initial calculation of the score function, though the value is zeroed out before we compute the probability.

Since the Analyzing Balance Changes (ABC-Meta) task attempts to approximate the true meta, we restrict ourselves only to the information to which a human player would have access. Specifically, these are pickrates, BST, Type Values and Meta Type Values. We do not use winrates or popularity as Pok\'emon Showdown does not track winrates and the popularity measure is not easily accessible. We run a grid search (Appendix \ref{appendix_gridsearch}) over these four components using the metrics defined in the Section \ref{meta_discovery_metrics} to determine our final team-building algorithm. Our final score function for the ABC-Meta task is:

\begin{equation}
    \text{Score}_{ABC Meta} = \text{Pickrates} \times \text{BST}
\end{equation}

For the BSD-Meta task where we have no prior knowledge of the metagame, we simulate the meta from scratch. We thus use both the winrates and popularity as a replacement for prior knowledge of the meta. 
In addition, we do not use the pickrates. If we used the pickrates, then the team-building algorithm would depend on the pickrates and the pickrates (in this scenario) would depend on the team-building algorithm. This could cause an undesirable feedback loop which would result in a small set of Pok\'emon constantly being selected. This is not ideal as the Pok\'emon picked initially at random could significantly affect the final meta. Using the winrate instead breaks this feedback loop while also allowing the algorithm to prioritize stronger picks.  
Our score function is also different from the ABC-Meta task in that we use a sum instead of a product to arrive at the overall score. We make this change because we want to encourage diversity in the meta. Due to the addition of the popularity term, multiplying all three quantities would generally result in smaller values. The exceptions would be those Pok\'emon with a high value in multiple terms. This poses a problem as it would result in the use of a smaller set of Pok\'emon without sufficient exploration of the others. Thus we use a sum to increase the effect of each individual term. 
Our final score function for the BSD-Meta task is:

\begin{multline}
\text{Score}_{BSD Meta} = \text{Winrates} + \text{a} \times \text{BST} 
+ \text{b} \times \text{Meta Type Value} \\
 + \text{c} \times \text{Type Value} + \text{Popularity}
\end{multline}

\noindent
Where $a$, $b$, and $c$ are constants that we set to 0.50, 0.25, and 0.25 respectively based on preliminary results.

For both the ABC-Meta and BSD-Meta tasks, we utilize an epsilon-greedy policy to pick Pok\'emon. We greedily select a Pok\'emon based on the scores derived above, but there is an epsilon chance of selecting a Pok\'emon using the inverse pickrate, that is, selecting a less frequently used Pok\'emon.
This introduces an element of exploration in the team-building system that ensures that the system does not always generate the same set of teams. In the case of the ABC-Meta where we already have an idea of the meta, our epsilon value is fixed at 0.001. When calculated over the 12 Pok\'emon in a battle, this means there is an approximately 1\% chance of picking a Pok\'emon based on the inverse pickrate. For the BSD-Meta where both the winrates and the popularity start with a default value of 0, we set the epsilon value to 1.0 and linearly decay it down to 0.001 over 20,000 battles. This allows for an initialization process where the system can obtain statistics for every legal Pok\'emon.

\section{Meta Discovery Environment}\label{meta_environment}
The final component of our meta discovery framework is the simulator environment. This environment must simulate battles using the battle agent, generate teams using our team-building system and also track statistics such as pickrates and winrates. 

Our agents are setup to be able to run multiple battles concurrently. This may cause a race condition if we update our battle statistics at the end of each battle. Instead, we perform this update at regular intervals. To help reduce technical overhead, we also elect to generate a large number of teams at this point and simply sample a team at the start of a battle. The number of teams generated and the interval at which we update battle statistics is a parameter. We use an arbitrary value of 2500 generated teams and an update frequency of 1000 battles. 

Since our goal is to approximate the metagame on Pok\'emon Showdown, we must follow its system of updating tiers every three months. Hence, we determine the number of battles needed to simulate for one month. The most popular non-random competitive metagame on Pok\'emon Showdown is the OU ladder \cite{smogon-stats}. This averages to about 1.45 million battles a month. The second most popular metagame is comparatively dwarfed at an average of 130,000 battles for the Ubers tier. We elect to use a fixed size number of battles per month for simplicity and chose 150,000 battles per month. In total, we simulate 3 months worth of changes for each battle and thus we perform 450,000 battles (roughly 8-14 hours). 
In addition, while presence in a tier is determined by usage rates, in practice this means 34 to 40 Pok\'emon are officially classified as a member of a tier. Thus we define the meta as the top 40 Pok\'emon in a tier. 

Finally, we also implement a blanket ban of the Little Cup (LC) tier in our system. That is, the Pok\'emon in the LC tier are not eligible to be picked by the team-building algorithm. We do this for two reasons. First, the Pok\'emon in the LC tier are extremely weak in relation to the rest of the Pok\'emon and can thus be easily identified. Second, they make up over a quarter of all the available Pok\'emon (210 out of 740). As a result, our team-building algorithm would spend unnecessary time simulating battles between these Pok\'emon. 

\section{Meta Discovery}\label{meta_discovery}

We now have all three components of our meta discovery system. We have two battle agents, the self-play PPO agent and the heuristic agent, a team-building algorithm for both our tasks, and a simulator environment to run meta discovery.

Our objective in the ABC-Meta task to analyze the effects on the metagame after a change, specifically a Pok\'emon being banned. This is a common occurrence in Pok\'emon Showdown \cite{smogon-tiering-policy} with such bans occurring every few months. To better evaluate our work, we specifically target instances where only a single ban is applied. In each scenario, we take the weighted usage statistics over the three month period immediately preceding the ban and average them with an equal weightage. We use this as the initial metagame, that is, the 40 most popular Pok\'emon in this list are classified as the meta Pok\'emon. We additionally also acquire the three month period after the ban to compare our framework against. In an ideal scenario, our discovered metagame would significantly overlap with this true metagame. 
We note that in general bans on Pok\'emon Showdown are enacted in the middle of a month. Since only statistics for the entire month are available, we do not consider the month in which the Pok\'emon is banned. Rather, we take the three months before the ban and the three months after the ban. 

To show that our system is not suited to just one metagame, we identify four scenarios spanning four tiers - OU, UU, NU, and PU. In each case, the initial metagame is relatively stable with no major changes to the meta in a 3-month period. Following this period, a single Pok\'emon is banned, thus causing a shift in the metagame. We identify these scenarios by crawling through the Smogon forums. We note that these bans are not available in a structured format and identifying any changes requires significant manual effort.
Table \ref{tab:meta-discovery-eval-scenarios} depicts the four scenarios that we consider in terms of the tier, Pok\'emon banned and the month of the ban. 

In the BSD-Meta task, we have no prior metagame. Rather, our aim is to analyze the discovered metagame starting from a blank slate. In an ideal scenario, the discovered metagame should primarily contain the strongest Pok\'emon, that is, the Pok\'emon in the highest tiers of Smogon (Anything Goes, Ubers, OU). 
Here, we remove all bans (including the ban we impose on LC) and simulate three months of battle. We thus have only one scenario to consider which we call ``Blank Slate".
This represents a baseline for the ABC-Meta task, but also a distinct problem with its own challenges.

\begin{table}[t]
\centering
\caption{The 4 scenarios we consider for evaluating the AST-Meta task.}
\label{tab:meta-discovery-eval-scenarios}
\begin{tabular}{|l|l|l|l|}
\hline
\textbf{Scenario} & \textbf{Tier} & \textbf{Ban} & \textbf{Month} \\
\hline
Smogon OU        & OU            & Kyurem        & 2021-12        \\
Smogon UU        & UU            & Aegislash     & 2022-06        \\
Smogon NU        & NU            & Blastoise     & 2022-09        \\
Smogon PU        & PU            & Vanilluxe     & 2022-08        \\
\hline
\end{tabular}
\end{table}

\subsection{Evaluation Metrics}\label{meta_discovery_metrics}
To evaluate our meta discovery framework on the ABC-Meta task, we develop two different metrics. Let us call the pre-ban metagame $A$, the post-ban metagame $B$, and our discovered metagame $B'$. 

Our first metric, Overlap, considers the overlap between $B$ and $B'$. This tells us how close the two metagames are by comparing the specific Pok\'emon in the meta. A value of 0\% would mean that none of the Pok\'emon in our discovered metagame $B'$ are present in $B$ while a value of 100\% would mean that all the Pok\'emon in $B'$ are present in $B$.

However, just identifying whether Pok\'emon are present in a metagame is not sufficient to fully evaluate our framework. The relative ordering of the Pok\'emon is also important. More specifically, we must compare the change in ranking of Pok\'emon from $A$ to $B$ compared to the change in ranking of Pok\'emon from $A$ to $B'$. We do this using the Edit Distance between the Pok\'emon in the two metagames. We calculate the Edit Distance as the mean absolute difference in ranking for all Pok\'emon present in both $A$ and $B$. This value tells us how much of a shift has occurred in the rankings of the Pok\'emon as a result of the change in meta. We can then compute the Edit Distance between $A$ and $B'$ and compare the two quantities. We note that even when the two quantities are close, it only means that the two metas have had a similar amount of change. It does not tell us that the same changes have occurred. To understand if these changes are correlated, we use Spearman's Rho, a statistical measure that measures rank correlation. A positive correlation indicates the degree to which the direction of changes to a Pok\'emon's ranking in $B$ matches the changes to the same Pok\'emon's ranking in $B'$ while a negative correlation indicates the inverse. The magnitude of the correlation indicates the strength of this relationship.

For the BSD-Meta task, we cannot use both the Overlap \& Edit Distance metrics from above as the pre-ban metagame $A$ does not exist. Instead, we analyze the discovered meta by comparing the distribution of Pok\'emon in $B'$ to the tiers on Smogon at the time of analysis (October 2022). Specifically, we compute the percentage of the top 3 Smogon tiers that have been captured by the discovered metagame. We use the top 3 tiers as they consist of approximately 80 Pok\'emon (Anything Goes has only 2 Pok\'emon). Thus we expect the majority of, if not the entire, discovered metagame to be within these tiers.

\subsection{Baselines}\label{meta_discovery_approaches}
Prior approaches cannot be used as baselines in this case as they either focus on other games \cite{league-champ-reco,Summerville_Cook_Steenhuisen_2021,pogo-counter,pogo-optimization}, require human interaction \cite{rejim-time-efficient}, significant reproduction effort \cite{technical-machine}, or lack public detail on the algorithm \cite{fsai}.

For the ABC-Meta tasks, we utilize two baselines. The first is what we term a naive baseline - we pretend the Pok\'emon that was banned doesn't exist anymore while everything else stays the same. That is, we take the usage statistics (pickrate) for the month preceding a Pok\'emon's ban and simply remove that Pok\'emon from the list. Thus all the Pok\'emon with lower usage than the banned Pok\'emon would be shifted up by one. 
For our other baseline, we use the same system we use for the BSD-Meta (using the Simple Agent). We expect this approach to perform poorly as it does not incorporate prior knowledge of the metagame, but we include it to demonstrate the importance of this knowledge. 
Since we use two agents as discussed in Section \ref{meta_battle}, we have two versions of our meta discovery framework. One uses the self-play PPO Agent and the other uses the heuristic Agent.

For the BSD-Meta task, there are no existing baselines due to the lack of prior work. Instead, we elect to craft a baseline that is just a sorted list of Pok\'emon based on their BST, as the BST plays a large role in determining a Pok\'emon's strength. 
We utilize the same two versions of our meta discovery framework as in the ABC-Meta task.

\subsection{Results}\label{meta_discovery_results}

\begin{table*}[t]
\centering
\caption{Evaluation of the meta discovery framework using the Overlap metric on the four ABC-Meta scenarios. Higher is better, 1.00 is perfect. \textbf{Bold} values indicate the best methods for each tier.}
\label{tab:meta-discovery-eval-overlap}
\begin{tabular}{|l|r|r|r|r|r|}
\hline
\textbf{Scenario} & \textbf{Blank Slate} & \textbf{Naive} & \textbf{Meta Discovery} & \textbf{Meta Discovery} \\
\textbf{} & \textbf{Discovery} & \textbf{Baseline} & \textbf{RL Agent} & \textbf{Heuristic Agent}\\
\hline
\textbf{Smogon OU} & 0.350 & 0.950 & \textbf{0.975} & \textbf{0.975} \\
\textbf{Smogon UU} & 0.375 & 0.975 & \textbf{1.000} & \textbf{1.000} \\
\textbf{Smogon NU} & 0.350 & 0.925 & \textbf{0.950} & \textbf{0.950} \\
\textbf{Smogon PU} & 0.300 & \textbf{0.925} & \textbf{0.925} & \textbf{0.925} \\
\hline
\end{tabular}
\end{table*}

\begin{table*}[t]
\centering
\caption{Evaluation of the meta discovery framework using the Edit Distance metric on the four ABC-Meta scenarios. Values indicate the delta from the True Meta. Lower is better, 0.00 is perfect. \textbf{Bold} values indicate the best methods for each tier.}
\label{tab:meta-discovery-eval-edit}
\begin{tabular}{|l|r|r|r|r|r|}
\hline
\textbf{Scenario} & \textbf{Blank Slate} & \textbf{Naive} & \textbf{Meta Discovery} & \textbf{Meta Discovery} \\
\textbf{} & \textbf{Discovery} & \textbf{Baseline} & \textbf{RL Agent} & \textbf{Heuristic Agent} \\
\hline
\textbf{Smogon OU} & 108.77 & 3.04 & \textbf{1.03} & 1.09 \\
\textbf{Smogon UU} & 82.48 & 4.60 & \textbf{1.30} & 1.50 \\
\textbf{Smogon NU} & 79.85 & 5.57 & 2.74 & \textbf{2.49} \\
\textbf{Smogon PU} & 77.74 & 6.84 & 0.33 & \textbf{0.11} \\
\hline
\end{tabular}
\end{table*}

Table \ref{tab:meta-discovery-eval-overlap} includes the results on the Overlap metric. 
We see that in all four cases, our meta discovery approach meets or outperforms the baselines. In three of four cases, the discovered meta is only a few Pok\'emon away from the true meta with a perfect result in the final case. The naive baseline performs well since the bans caused small shifts in the meta that moved a few Pok\'emon in and out of the meta. The Blank Slate Discovery method performs the worst, achieving only about 35\% Overlap in most cases. This could be due to it having no prior knowledge of the metagame. 

Table \ref{tab:meta-discovery-eval-edit} depicts the results of our Edit Distance metric. Specifically, we show the delta from the True Meta. Even here, the Blank Slate Discovery baseline is significantly worse than the others, having discovered a vastly different metagame. The naive baseline is better since it effectively just shifts some percentage of the meta ranking up by one ranking. However, our meta discovery approach once again performs the best, with Edit Distances that are very close to the true meta.

Looking at the results of our Meta Discovery approach between the RL and heuristic agents, we see that there is no significant difference in terms of their overall metrics. This is to be expected to some extent as both agents were quite close to each other in terms of overall performance. 

Considering both metrics together, we see that though the naive baseline has a high level of overlap, it does not account for the shifts within the meta itself. Further, the Blank Slate Discovery baseline suffers heavily due to the lack of metagame knowledge. Our approach on the other hand has a greater overlap while also simulating the shifts within the metagame.
We note that despite outperforming the baselines, our Meta Discovery approach is still some way from the True Meta. This shows that there is room for improvement in our team-building approach \& trained agents. Despite this, we present clear evidence that our approach is able to approximate the effects of a change in the meta to a high degree.

\begin{table}[t]
\centering
\caption{Results of Spearman's Rho on the Meta Discovery approaches. }
\label{tab:meta-discovery-eval-corr}
\begin{tabular}{|l|l|r|r|}
\hline
\textbf{Scenario} & \textbf{Agent} & \textbf{Rho} & \textbf{p-value} \\ \hline
Smogon OU         & RL         & 0.1861       & 0.2502           \\ \hline
Smogon OU         & Heuristic      & 0.1827       & 0.2589           \\ \hline
Smogon UU         & RL         & -0.1455      & 0.3702           \\ \hline
Smogon UU         & Heuristic      & -0.1304      & 0.4224           \\ \hline
Smogon NU         & RL         & -0.0645      & 0.6960           \\ \hline
Smogon NU         & Heuristic      & -0.0686      & 0.6779           \\ \hline
Smogon PU         & RL         & 0.1152       & 0.4787           \\ \hline
Smogon PU         & Heuristic      & 0.1482       & 0.3613           \\ \hline
\end{tabular}
\end{table}

As discussed in Section \ref{meta_discovery_metrics}, the Edit Distance only tells us how similar the quantity of changes to the meta are, not how close the two metas are. To study this, we compute Spearman's Rho over our meta discovery approaches in Table \ref{tab:meta-discovery-eval-corr}. We see that the two agents are once again quite similar in terms of results.
Looking at the rho value, the correlation in all cases is weak and varies between positive and negative. 
However, the high p-value indicates that there is no statistically significant relationship between the edit distances of the true meta and our discovered metas.
This could mean that our discovered meta, despite it's similarity to the true meta in terms of overlap and edit distances, is still a distinct metagame. 
That is, while the Pok\'emon used are largely the same (at least 92.25\% overlap at a minimum as seen in Table \ref{tab:meta-discovery-eval-overlap}, their relative usage in the discovered metagame is different.
Although this is not ideal, it serves to re-emphasize how difficult this problem of predicting a Pok\'emon's placement in the meta is, especially without human expertise guiding it.

\begin{table}[t]
\centering
\caption{Evaluation on the BSD-Meta task. Values are the percentage of the Smogon tiers in the discovered meta.}
\label{tab:meta-discovery-eval-bsd}
\begin{tabular}{|l|r|r|r|}
\hline
\textbf{Agent}  & \textbf{AG} & \textbf{Ubers} & \textbf{OU} \\ \hline
BST Baseline    & 100\%       & 71.79\%        & 22.86\%     \\ \hline
Simple Agent    & 100\%       & 61.54\%        & 17.14\%     \\ \hline
Heuristic Agent & 100\%       & 64.10\%        & 17.14\%     \\ \hline
\end{tabular}
\end{table}

\begin{table}[t]
\centering
\caption{Composition of the metas discovered in the BSD-Meta task. Values are the percentage of the discovered meta that is classified in a particular Smogon tier.}
\label{tab:meta-discovery-eval-bsd2}
\begin{tabular}{|l|r|r|r|r|}
\hline
\textbf{Agent}  & \textbf{AG} & \textbf{Ubers} & \textbf{OU} & \textbf{Below OU} \\ \hline
BST Baseline    & 5.0\%       & 70.0\%         & 20.0\%      & 5.0\%             \\ \hline
Simple Agent    & 5.0\%       & 60.0\%         & 15.0\%      & 20.0\%            \\ \hline
Heuristic Agent & 5.0\%       & 62.5\%         & 15.0\%      & 17.5\%            \\ \hline
\end{tabular}
\end{table}

Table \ref{tab:meta-discovery-eval-bsd} depicts the results of our system on the BSD-Meta task. As discussed previously, we analyze the percentage of the top three Smogon tiers (AG, Ubers, OU) that our system has predicted as being in the meta. In all cases, the Anything Goes tier is captured in its entirety and over 60\% of the Ubers tier is captured. However, only around 20\% of the Pok\'emon in the OU tier are captured, rightly indicating that there's a sizable gap in the strength between Poke\'emon in OU and Ubers. 
In Table \ref{tab:meta-discovery-eval-bsd2}, we instead examine the composition of our discovered meta in terms of the Smogon tiers. That is, we examine how much representation each of the Smogon tiers have in our discovered meta. In all cases, at least 80\% of the discovered meta is within the top 3 Smogon tiers. However, our framework performs worse than the baseline, capturing more Pok\'emon in lower tiers. We suspect that this happens for two reasons. First, our framework might have found Pok\'emon that work well as counterpicks to the meta but may not be very strong otherwise. Second, our framework might be exploring too much and not exploiting enough. In standard Smogon tiers, 50-80\% of picks are from the meta, that is, the top 40 most popular Pok\'emon. We observed the same effect in our ABC-Meta task results. However, in the case of the BSD-Meta task where we don't have prior knowledge of the meta, we are forced to explore more often. We found that the top 50\% of picks in the discovered meta span across 300+ Pok\'emon. This tells us that we need a more complex system to pick Pok\'emon for our teams. 
A solution here would be modifying the team-builder's score function for the BSD-Meta (exploration-focused) to slowly move towards the ABC-Meta (exploitation-focused) as time goes on. We leave this task for future work.
On comparing our agents, the results are quite close, with the heuristic agent's metagame picking one Pok\'emon more in the Ubers tier. As discussed earlier, this is expected due to the similarity in performance between the agents. 
Overall, we view these results as indicating a significant utility of our approach to game designers in terms of helping to predict the impact of balance changes by approximating the behavior of a player base.

\section{Future Work}\label{meta_future}

Although our work is implemented specifically on Pok\'emon Showdown, it presents a framework for a general purpose simulation of the metagame. We believe it offers game designers an avenue to better understand how changes to game balance may affect the meta.  
We believe our framework can be extended to examine the short- or long-term affects of balance changes, not just a static time period. For instance, instead of analyzing the resultant meta after the equivalent of 3-months of games, we might examine what it looks like after a single day, a single month, or an entire year.
In addition, we are generating statistics such as the pickrate and winrate for individual characters as a byproduct of simulating the metagame. This confers the advantage of understanding the impact balance changes have on specific characters. 
We anticipate that we could further vary the degree of balance change from the single character ban presented in this paper. 
For instance, multiple new characters could be added or banned, character attributes could be modified, entirely new mechanics could be added etc. 

There are also improvements that can be made to individual components of our system. As discussed in section \ref{meta_battle}, an ideal scenario would be to have a swarm of agents of varying skill levels to better emulate the playerbase. This would naturally require agents that are both stronger and weaker than the agents used in this paper. One way to achieve this would be to train an agent in a manner similar to \cite{huang-selfplay}, but clone it at different stages of the training process to act as players of different skill levels. In addition, we may want agents to employ different styles of play - one may want to do as much damage as possible each turn while another might take a more conservative approach and attempt to preserve their own team while dealing a little damage each turn.

Similarly, on the team-building side, building teams tailored to agent playstyles is of particular interest. This is a harder task and would either require manual input of domain knowledge, or a machine learning agent responsible for building teams. In such a scenario, it could be worth exploring a system where both the battle agent and the team-builder are trained together - with each battle simulated potentially improving both the team-builder and the battle agent.

Overall, we anticipate our framework could serve as a general purpose system to simulate a metagame given a team-building algorithm and a game-playing agent. 
One exception is the introduction of entirely new systems. For instance, if a drafting phase (as analyzed in \cite{Summerville_Cook_Steenhuisen_2021}) was incorporated, we would need an additional component to handle this and interface with the team-builder.

One special case involves games like League of Legends with multiple human players on each team, each with different responsibilities. The simulation here would be more complex as each player would require an agent trained for their particular role. However, the rest of our framework would largely be the same in terms of simulating matches and tracking statistics.

\section{Conclusion}\label{meta_conclusion}
In this paper, we introduced a framework for evaluating changes to game balance via meta discovery. We developed a Reinforcement Learning agent to approximate human players and created a team-generation system that adapts to changes in the meta. We implemented this system in the Pok\'emon Showdown competitive online game and evaluated the effectiveness of this system on two separate tasks. Our results show that the framework is effective in approximating the results of balance changes to the metagame.

\subsection{Acknowledgments}
\noindent This work was funded by the Canada CIFAR AI Chairs Program, Alberta Machine Intelligence Institute, and the Natural Sciences and Engineering Research Council of Canada (NSERC).

\bibliographystyle{IEEEtran}
\bibliography{IEEEabrv,main}

\newpage
\appendices
\section{Team-building Algorithm: Grid Search Results}\label{appendix_gridsearch}

As discussed in Section \ref{meta_teambuilding_algos}, we perform a grid search across 4 components to determine the best team-building algorithm for the ABC-Meta task. These are the pickrate, the BST, The Meta Type Value and the Type Value. We use the following basic algorithm:

\begin{equation}
\begin{split}
    \text{Score}_{ABC Meta} = \text{Pickrate} \times (\text{c1} \times \text{BST} + \\ \text{c2} \times \text{Meta Type Value} + \text{c3} \times \text{Type Value})
\end{split}
\end{equation}

This function has two component parts. The first term, the pickrates, allows us to use the current metagame statistics to impact our team selection. These pickrates range from a value of 0 to 1, with a value of 1 indicating that every team uses a particular Pok\'emon. We weigh this by our second term, which combines domain knowledge (BST), meta knowledge (Meta Type Value) and team-building principles (Type Value). This second term also ranges from 0 to 1, but each sub-term is weighted differently. 

We try out 8 combinations of weights for the three sub-terms, including a base case where we remove them entirely as well as scenarios where we try them out in isolation. For our last scenario where all three of them are tried in combination, we use weights of $0.50$, $0.25$, and $0.25$.
Since the BST is a strong indicator of a Pok\'emon's strength, we weigh it higher. The two type scores do not consider the strength of a Pok\'emon, just their types. In addition, since they operate in similar areas, we weigh them equally. While we experimented with additional weighing schemes, our preliminary results indicated that these worked best.

For the purposes of this grid search, we use the RL agent and evaluate using the same metrics discussed in Section \ref{meta_discovery_metrics}. We consider only the Smogon OU scenario from Section \ref{meta_discovery} and follow the same experimental setup described in that section. 
We fix the usage of the pickrates and toggle the use of the remaining three components in all combinations. In total, we compare eight approaches. Table \ref{tab:appendix-gridsearch} depicts these results.

\begin{table}[ht]
\centering
\caption{Results of performing a grid search over the three domain knowledge components of our team-building algorithm.}
\label{tab:appendix-gridsearch}
\begin{tabular}{|l|l|l|l|l|}
\hline
\textbf{c1} & \textbf{c2} & \textbf{c3} & \textbf{Edit Distance} & \textbf{Overlap} \\ \hline
0   & 0    & 0    & 2.83  & 95.0\% \\ \hline
1   & 0    & 0    & 1.09  & 97.5\% \\ \hline
0   & 1    & 0    & 8.64  & 82.5\% \\ \hline
0   & 0    & 1    & 0.18  & 92.5\% \\ \hline
0.5 & 0.5  & 0    & 1.82  & 92.5\% \\ \hline
0.5 & 0    & 0.5  & 1.58  & 92.5\% \\ \hline
0   & 0.5  & 0.5  & 2.33  & 90.0\% \\ \hline
0.5 & 0.25 & 0.25 & 0.85  & 92.5\% \\ \hline
\end{tabular}
\end{table}

Examining the results, we see that the use of the BST has a noticeable impact on the Overlap, achieving a minimum of 92.5\% Overlap and around 1 in Edit Distances. 
Using the Meta Type Value on the other hand has a significant negative impact on both metrics, with an Overlap of only 82.5\% when used in isolation. This seems to be true even when it is used in conjunction with the other components, with the versions using the Meta Type Value performing worse than the ones that do not. 
Using only the Type Values has the closest Edit Distance, but does not accurately predict the meta in terms of overlap. Using only the pickrate (the first row) achieves slightly higher overlap at the cost of a much higher Edit Distance. Overall, the version which uses only the pickrate and the BST seems the best choice. It has the highest Overlap in the table and does reasonably well in terms of Edit Distances.

\end{document}